\documentclass[conference]{IEEEtran}
\IEEEoverridecommandlockouts
\usepackage{cite}
\usepackage{amsmath,amssymb,amsfonts}
\usepackage{algorithmic}
\usepackage{graphicx}
\usepackage{url} 
\usepackage{textcomp}
\usepackage{xcolor}
\def\BibTeX{{\rm B\kern-.05em{\sc i\kern-.025em b}\kern-.08em
    T\kern-.1667em\lower.7ex\hbox{E}\kern-.125emX}}

\usepackage{tikz}
\usepackage{textcomp}
\usepackage{hyperref}
\usepackage{lipsum}

\newcommand\copyrighttext{%
  \footnotesize \textcopyright 2025 IEEE. Personal use of this material is permitted.
  Permission from IEEE must be obtained for all other uses, in any current or future
  media, including reprinting/republishing this material for advertising or promotional
  purposes, creating new collective works, for resale or redistribution to servers or
  lists, or reuse of any copyrighted component of this work in other works.}
\newcommand\copyrightnotice{%
\begin{tikzpicture}[remember picture,overlay]
\node[anchor=south,yshift=10pt] at (current page.south) {\fbox{\parbox{\dimexpr\textwidth-\fboxsep-\fboxrule\relax}{\copyrighttext}}};
\end{tikzpicture}%
}
    
\begin{document}

\title{SteelBlastQC: Shot-blasted Steel Surface Dataset with Interpretable Detection of Surface Defects
}

\author{\IEEEauthorblockN{Irina Ruzavina, Lisa Sophie Theis, Jesse Lemeer, Rutger de Groen, Leo Ebeling, Andrej Hulak, Jouaria Ali,\\ Guangzhi Tang, Rico Mockel}
\IEEEauthorblockA{\textit{Department of Advanced Computing Sciences, Maastricht University, Maastricht, The Netherlands} \\
\{guangzhi.tang, rico.mockel\}@maastrichtuniversity.nl}}

\maketitle
\copyrightnotice

\begin{abstract}
Automating the quality control of shot-blasted steel surfaces is crucial for improving manufacturing efficiency and consistency. This study presents a  dataset of 1654 labeled RGB images (512$\times$512) of steel surfaces, classified as either “ready for paint” or “needs shot-blasting.” The dataset captures real-world surface defects, including discoloration, welding lines, scratches and corrosion, making it well-suited for training computer vision models. Additionally, three classification approaches were evaluated: Compact Convolutional Transformers (CCT), Support Vector Machines (SVM) with ResNet-50 feature extraction, and a Convolutional Autoencoder (CAE). The supervised methods (CCT and SVM) achieve 95\% classification accuracy on the test set, with CCT leveraging transformer-based attention mechanisms and SVM offering a computationally efficient alternative. The CAE approach, while less effective, establishes a baseline for unsupervised quality control. We present interpretable decision-making by all three neural networks, allowing industry users to visually pinpoint problematic regions and understand the model’s rationale. By releasing the dataset and baseline codes, this work aims to support further research in defect detection, advance the development of interpretable computer vision models for quality control, and encourage the adoption of automated inspection systems in industrial applications.
\end{abstract}


\section{Introduction}

During the manufacturing and assembly of steel products, surface texture defects are common and can significantly impact the quality and appearance of the finished product, especially during subsequent processes like painting \cite{Scott1993}. Such defects include rust, scratches, and untreated or insufficiently treated areas (particularly along welding seams), characterized by smooth and shiny surfaces instead of the desired matte finish. Shot-blasting is a widely used method to address these defects, where steel particles are projected at high speeds onto the surface, creating a uniform and rough texture that prepares the material for further processing \cite{Hansel2000}. This process is essential for improving surface quality and ensuring that products meet the required standards. However, the efficiency of traditional shot-blasting is often limited by human factors, as workers face significant health and safety risks, including exposure to toxic dust, high noise levels, and heat stress. These hazards restrict productivity to approximately 15–20m² per hour \cite{Nguyen2024}. Automating the shot-blasting process can enhance both productivity and consistency by establishing objective performance standards, while also reducing the need for manual labor and minimizing long-term operational costs \cite{Holzknecht2005}.

The quality control of shot-blasted surfaces is a crucial step within this process. Usually, experienced operators visually assess the surface texture as they treat the steel. To replace the manual quality control with enough confidence to keep the high accuracy and efficiency, an advanced automated approach is needed. Computer Vision (CV) methods, especially deep learning models, can capture both fine-grained details like scratches and broader texture patterns, ensuring reliable classification. Other methods, such as a contact sensor, that measure the texture of the surface, would either be too inefficient when using a high resolution or too inaccurate when measuring bigger parts of the surface at once. Since both global and local characteristics of the metal play a role when classifying surface parts to be ready for paint or not, CV approaches are an ideal choice \cite{Holzknecht2005}.

For CV approaches to be effective in an industrial setting, the choice of the dataset to train the models on is a key consideration. Many widely used datasets represent idealized or limited defect types \cite{prunella2023deep}, whereas a purpose-built dataset can contain a diverse range of anomalies and surface textures encountered in real-world industrial settings for shot-blasting, which can improve the performance of tailored CV models. Furthermore, integrating interpretable deep-learning approaches is vital for industry adoption \cite{li2022interpretable}, as transparency in model decision-making builds trust among quality control professionals. Interpretability ensures compliance with industry standards and safety regulations, ultimately paving the way for more reliable and accountable deployment in production environments.

In this paper, we introduce SteelBlastQC\footnote{The proposed open-source SteelBlastQC dataset is hosted on DataverseNL (https://doi.org/10.34894/EKZNN0), baseline codes are available on GitHub (https://github.com/ERNIS-LAB/SteelBlastQC)}, a realistic dataset comprising 1654 high-resolution RGB images (512$\times$512 pixels) of shot-blasted steel surfaces collected directly from an industrial manufacturing facility. Each image is labeled by domain experts as either “ready for paint” or “needs shot-blasting,” reflecting practical quality control standards. To enhance generalizability, SteelBlastQC includes a wide range of surface defects and variations in shot-blasted finishes. In addition, we benchmark both supervised and unsupervised deep learning algorithms on this dataset, including Compact Convolutional Transformers (CCT) \cite{CCTpaper}, Support Vector Machines (SVM) \cite{Vapnik1995} leveraging features extracted from ResNet-50 \cite{He2016}, and a Convolutional Autoencoder (CAE) \cite{cae}. Our experiments demonstrate that supervised approaches achieve up to 95\% classification accuracy, underscoring the dataset’s utility. Besides, by integrating interpretable deep-learning techniques, we analyze the regions that drive model decision-making, with the interpretation results effectively highlighting defect areas that justify the predictions.

\section{Related Works}

There has been a rise in CV methods in the identification of steel surface defects in manufacturing processes, especially in the last 5 years \cite{Ibrahim2024}. However, publicly available solutions are limited and not fully applicable to the task under our investigation.

One dataset published in 2016 that has been used in many publications is the NEU Metal Surface Defects Database, which collects six kinds of typical surface defects of steel strips: rolled-in scale, patches, crazing, pitted surface, inclusion, and scratches \cite{NEU_Metal_Surface_Defects}. This dataset has been used to train deep-learning models to classify metal surface defects \cite{siddiqui2024faster}. The focus here is on the classification of defects, not on surface texture. Other datasets that are more suitable for the task at hand are not available for free use. 
For instance, Texture-AD contains a diverse range of surface texture and defect data captured using advanced imaging techniques \cite{TextureAD2024}. This dataset has been specifically assembled to support the development of machine learning models for anomaly detection in industrial quality control, enabling the classification of more complex and subtle patterns. However, it is not freely available for use, which limits its accessibility for this study and broader research applications.

The NEU dataset has been used for the quality control of the surfaces of rolled products \cite{Benabdelkader2022}. For example, in this study, the authors aimed to develop an intelligent system to detect and classify surface defects in hot-rolled steel strips using a combination of convolutional neural network (CNN) and support vector machines (SVM). They used a modified version of the well-known AlexNet CNN for feature extraction and then used these features for the SVM classifier, to classify the six types of surface defects. This approach led to promising results, achieving an accuracy rate of 99.70\%. 

\section{SteelBlastQC Dataset}
\subsection{Dataset Description}
\begin{figure}
    \centering
    \includegraphics[width=0.45\textwidth]{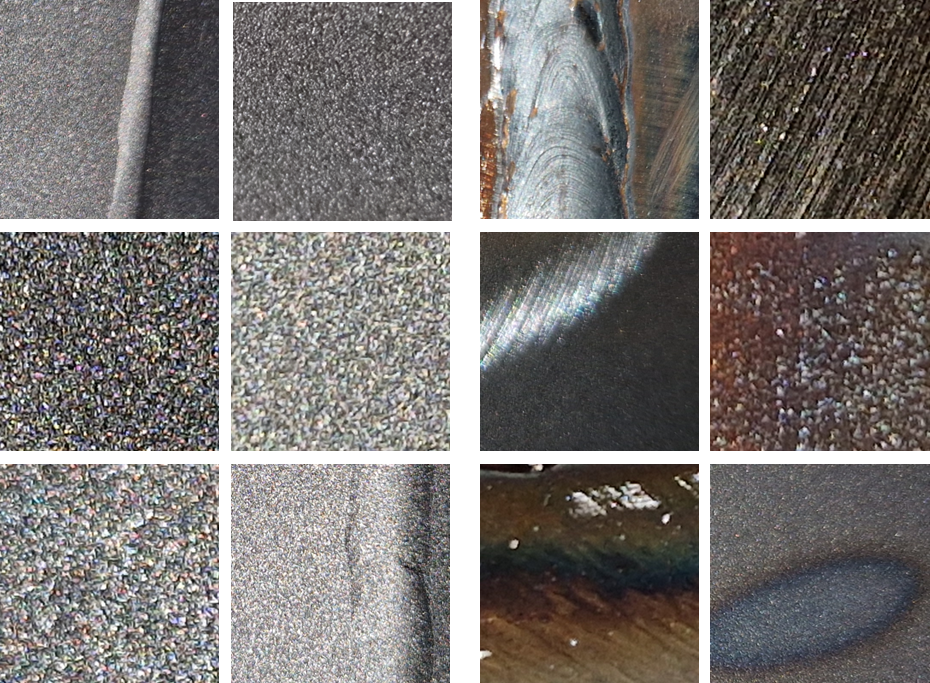} 

    \caption{Example of data collected: passing quality control (left) and failing quality control (right). The defects depicted on the right are (from top to bottom) a welding line, two types of scratches, corrosion, and two examples of discoloration.}
    \label{examples}
\end{figure}

We present the open-source SteelBlastQC dataset for the detection of surface defects for quality control of shot-blasted steel surfaces. The dataset consists of 1654 RGB images (512$\times$512) of steel surfaces that are either shot-blasted or still need shot-blasting to achieve the required texture, forming a binary classification task. The dataset includes 888 "ready for paint" images and 766 "needs shot-blasting" images. Examples can be found in Figure~\ref{examples}, with shot-blasted steel surfaces on the left and steel surfaces that need shot-blasting on the right. As declared by the collaborating manufacturer, the ideally treated surface is clean and uniformly coarse with an average roughness level of SA 2.5 \cite{ISO8501}.
Shown in the right half of Figure~\ref{examples}, are several types of defects to the surface, located by industrial shot-blasting experts. Areas with fresh welding lines and cuts clearly need shot-blasting, as the texture is visibly far from the required standard. Furthermore, defects such as abrasion, corrosion, and discoloration may occur due to mechanical or chemical exposure during the product assembly and must be treated. 

\subsection{Dataset Collection Process}
Images were collected from the outer part of a steel yacht hull, half of which was shot-blasted and verified by an expert and the other half was left untreated. We collected a similar proportion of "needs shot-blasting" and "ready for paint", aiming to avoid class imbalance. Since some untreated areas already have the desired texture, we aimed to focus on parts with obvious defects, such as discoloration, scratches, or welds. The camera used was the Fujifilm X-T30 II. The images were captured under controlled lighting and at a distance of 20-30 cm. 

\subsection{Dataset Preprocessing}
The original larger-size images were each cropped to a square of 2048$\times$2048 pixels at the center, which was then cut into 16 samples of size 512$\times$512 (see Figure~\ref{fig:architecture_overview}). This was done to ensure a consistent angle between the camera and the hull surface. The resulting set contained square RGB images that the models could handle quickly without losing texture-related information. Lastly, the data were split into a training and a testing set (ratio of 80/20), allowing for consistent training and fair performance assessment on unseen data out-of-the-box.

\section{Methods}
\label{sec:methods}

\begin{figure*}
    \centering
    \includegraphics[width=1\linewidth]{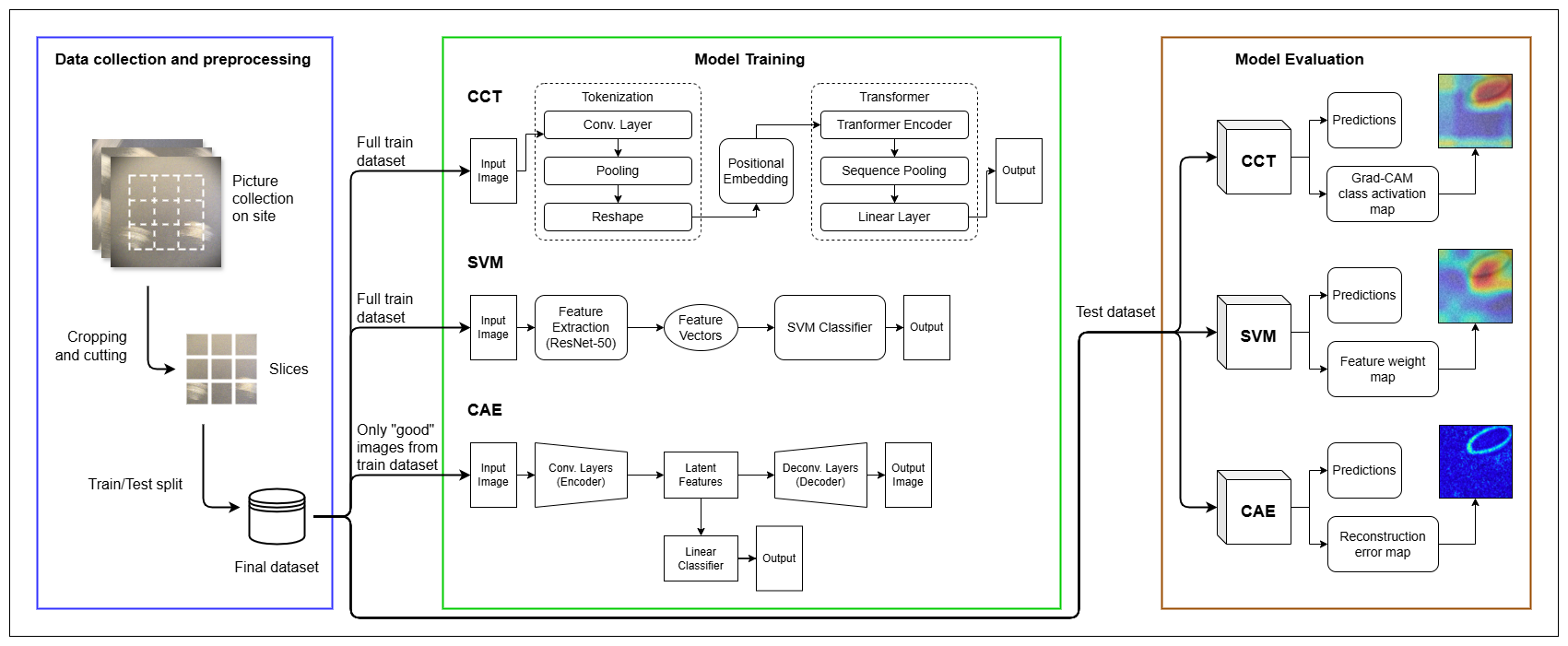}
    \caption{Our proposed interpretable surface defect detection pipeline using supervised and unsupervised deep learning approaches}
    \label{fig:architecture_overview}
\end{figure*}

We benchmarked the proposed dataset using both supervised and unsupervised deep learning approaches. Figure~\ref{fig:architecture_overview} shows the surface defect detection pipeline with SteelBlastQC, including dataset preprocessing, model training, the interpretable model evaluation on three different benchmarked methods.

\subsection{Support Vector Machine (SVM) with ResNet50 Feature Extraction}

Support Vector Machines combined with deep feature extraction provide a balance between computational efficiency and interpretability \cite{Vapnik1995}. While convolutional neural networks excel at feature learning, their decision processes often lack transparency. By employing ResNet-50 as a feature extractor and applying SVM for classification, this approach ensures robust defect detection while maintaining model interpretability.

ResNet-50, a deep residual network trained on the ImageNet dataset, is used to extract hierarchical texture features from steel surface images \cite{He2016}. The model’s final classification layer is removed, and a global average pooling layer is applied to generate 2048-dimensional feature vectors. Images are resized to 224$\times$224 pixels for consistency with ResNet-50’s architecture. To evaluate the impact of color information, both RGB and grayscale images are tested, with RGB images standardized using precomputed mean and standard deviation values.

Extracted feature vectors are fed into a linear SVM classifier, which constructs optimal hyperplanes for binary classification \cite{Vapnik1995}. A linear kernel with a regularization parameter $C=1.0$ is used to balance margin maximization and misclassification. This setup ensures computational efficiency while preserving the model’s ability to distinguish surface defects from paint-ready surfaces.

To interpret the SVM’s decision-making process, feature importance-based heatmaps are generated \cite{Selvaraju2017}. Unlike deep learning heatmaps, which rely on backpropagation-based attention mechanisms, SVM heatmaps are derived from the feature weights assigned to each extracted representation. The generated heatmaps are overlaid onto the original images using a colormap, highlighting regions that influenced classification.

\subsection{Compact Convolutional Transformer (CCT)}

Compact convolutional transformers enhance the vision transformer architecture \cite{CCTpaper} by integrating convolutional tokenization and sequence pooling to address challenges in traditional vision transformers. The model uses attention mechanisms to model long-range dependencies \cite{Attentionpaper}, augmented by the spatial inductive biases of convolutions. This hybrid architecture makes CCT well-suited for surface quality classification, where both local textures and global patterns are critical.

The convolutional tokenizer serves as the first stage of feature extraction, several convolutional layers to capture low-level spatial features like edges and textures. Subsequent pooling layers further compact the feature maps, reducing the spatial resolution. The resulting feature maps are flattened into a sequence of tokens. The transformer encoder, consisting of multiple layers with multi-head self-attention and feed-forward networks, processes these tokens to capture both local and global dependencies. 

Unlike traditional approaches that use class tokens, CCT employs sequence pooling to aggregate tokens into a single representation. This mechanism assigns attention-based importance weights, emphasizing critical features in the final compact representation. A fully connected layer then produces a probability distribution for binary classification. The architecture effectively combines convolutional inductive biases with transformer flexibility, yielding strong performance on small-scale datasets.

To interpret model predictions, Grad-CAM visualizations are implemented that highlight regions of interest influencing the classification \cite{GradCAM}. In contrast to the regular approach, which focuses on averaging the gradients globally over the spatial dimensions, our implementation focuses only on the last convolutional layer of the model's tokenizer. Forward and backward hooks were employed to capture activations and gradients without altering the model's structure. The heatmaps were normalized to the range $[0, 1]$ for consistent visualization and overlaid onto the original images using a colormap. Combined heatmaps were created by aggregating attention across spatial positions, offering a holistic view of the model’s focus. During evaluation on test data, the combined heatmaps were overlaid onto input images, showcasing areas like defects or scratches that guided the model's decisions.

\subsection{Convolutional Autoencoder (CAE)}

Convolutional Autoencoder (CAE) extends the basic Autoencoder (AE) architecture by using CNNs instead of dense layers, and is commonly used on image data \cite{cae}. The encoder contains a series of convolutional layers to compress the input into a low-dimensional representation, while the decoder performs deconvolution to generate a reconstruction \cite{threshold}. Then learned representation of the training data can then be compared with features of a test sample to determine whether it is defective or not. An important distinction of this approach is that it is fully unsupervised, meaning it does not require the training samples to have labels.

The explored configuration involved two distinct components: a Convolutional Autoencoder (CAE) for feature learning and reconstruction and a Statistical Process Control (SPC) threshold to classify the input images based on the reconstruction error, following \cite{threshold}. The encoder was comprised of three convolutional layers. Following the convolutional layers, the feature maps were flattened into a one-dimensional vector and passed through a fully connected layer to obtain a compact latent representation of 256 dimensions. The decoder was mirroring the encoder with the Sigmoid activation function applied to the output layer.

The described CAE was trained to minimize the reconstruction loss calculated as the Mean Squared Error (MSE), encouraging focus on pixel-level accuracy. Lambda regularization (\(\lambda\) =  0.1) was added to reduce variance in the learned features.

Then, the decision threshold was derived from the latent space statistics: 
\begin{equation}
T_d = \mu_d + C \cdot \sigma_d
\end{equation}
where  \(\mu_d\) is the mean of the distances, \(\sigma_d\) is the standard deviation of the distances and C is the constant term set to a low value (0.01) in order to prioritize recall. Finally, images producing error \(>T_d\) were classified as defective.

\section{Experiments and Results}

To ensure a fair comparison, all approaches were tested using a standardized experimental setup. Performance was evaluated using multiple metrics, including accuracy, F1-score, precision, and recall. Additionally, heatmaps were used visualization purposes. The results for each model are summarized in Table \ref{tab:method_comparison}.

\subsection{Support Vector Machines (SVM)}

\subsubsection{Experimental Setup}

The images in the RGB and grayscale dataset variants were resized to $224\times224$. Separate models were trained for both datasets to allow direct performance comparison. The implementation was carried out using Python and PyTorch. PyTorch was used for feature extraction with the pre-trained ResNet-50 model, and scikit-learn was employed for the SVM classifier.

\subsubsection{Results}

Training the model on the RGB dataset achieved higher accuracy of $94.5$\% compared to the grayscale dataset ($91.4$\%). This indicates that incorporating color information helps the model distinguish defects more effectively. The RGB version achieved an F1-score of $0.95$, with both recall and precision being identical, indicating a well-balanced classification performance.

\subsubsection{Heatmaps}

\begin{figure}[htbp]  
    \centering
    \includegraphics[width=0.45\textwidth]{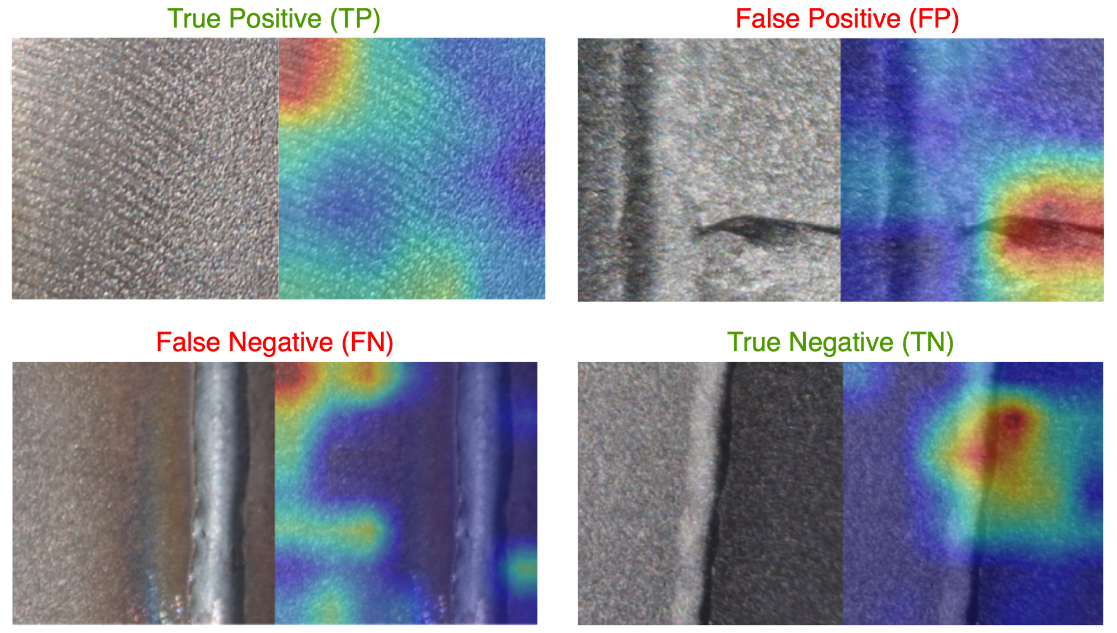}  
    \caption{Interpretation heatmap examples of truly and falsely classified images using SVM}
    \label{fig:svmfullheatmap}  
\end{figure}

We generated heatmaps to visualize the model’s focus areas during defect classification, highlighting the key regions that contributed to its predictions. Figure~\ref{fig:svmfullheatmap} showcases four different classification cases of SVM using RGB images: true positive (TP), false positive (FP), false negative (FN), and true negative (TN). 

In the true positive example, the model correctly identifies a defect, as seen in the heatmap where the activation strongly aligns with the actual defect location. The false positive case shows an incorrect classification, where the model mistakenly detects a defect in a non-defective region, indicated by heatmap activation in an irrelevant area. The false negative example illustrates a failure to detect an actual defect, despite some attention from the model in the heatmap. In contrast, the true negative case demonstrates a correct classification of a non-defective area, with the heatmap showing minimal activation in critical regions.

\begin{figure}[htbp]  
    \centering
    \includegraphics[width=0.45\textwidth]{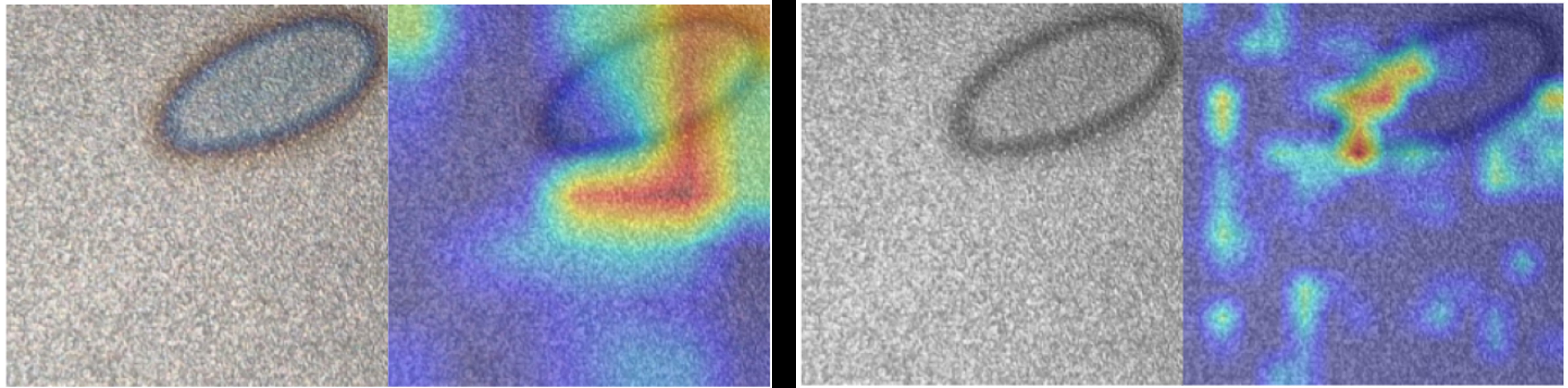}  
    \caption{Comparing interpretation heatmap of SVM using RGB (left) and grayscale (right) input image with defective surface}
    \label{fig:svmRGBGSheatmap}  
\end{figure}

While the RGB images achieved higher overall classification accuracy, grayscale heatmaps provided better interpretability. This is due to grayscale images emphasizing structural details more effectively, revealing how the model perceives textures and patterns. The variations in heatmap intensity offer insight into the decision-making process, helping to understand why certain misclassifications occur and where the model’s attention is directed during defect analysis. This result can also be seen in Figure~\ref{fig:svmRGBGSheatmap}.

\subsection{Compact Convolutional Transformers (CCT)}

\subsubsection{Experimental Setup}

The RGB images were resized to $128\times128$ pixels. Each model was trained for $20$ epochs using the AdamW optimizer with a learning rate of $0.001$ and a batch size of $32$. The implementation was conducted using Python. The CCT model was developed using PyTorch alongside the vit-pytorch library \cite{vitpytorch} for configuration and training.

\subsubsection{Results}

We experimented with different hyperparameters to determine the best balance between performance and computational efficiency. We selected $3$ model configurations to experiment with, depending on the number of parameters, that performed the best. The small CCT model consisted of $2$ convolutional layers, $4$ transformer layers, $4$ attention heads, and an embedding dimension of 64, with approximately $150,000$ parameters. This model achieved a test accuracy of $93.5$\%. The medium CCT model, with approximately $1$ million parameters, had $4$ convolutional layers, $8$ transformer layers, $8$ attention heads, and an embedding dimension of $256$. It achieved a test accuracy of $95$\%, offering the best trade-off between performance and efficiency. The large CCT model, containing $73$ million parameters, had $4$ convolutional layers, $12$ transformer layers, $12$ attention heads, and an embedding dimension of $768$. While this model achieved a comparable test accuracy of $94.5$\%, it required significantly more computational resources, making it less practical for deployment. 

The medium CCT model, which performed the best, achieved an F1-score of $0.95$, with recall and precision being the exact same. This suggests that the model effectively captures relevant features while maintaining consistency across the different metrics.

\subsubsection{Heatmaps}

\begin{figure}[htbp] 
    \centering
    \includegraphics[width=0.9\linewidth]{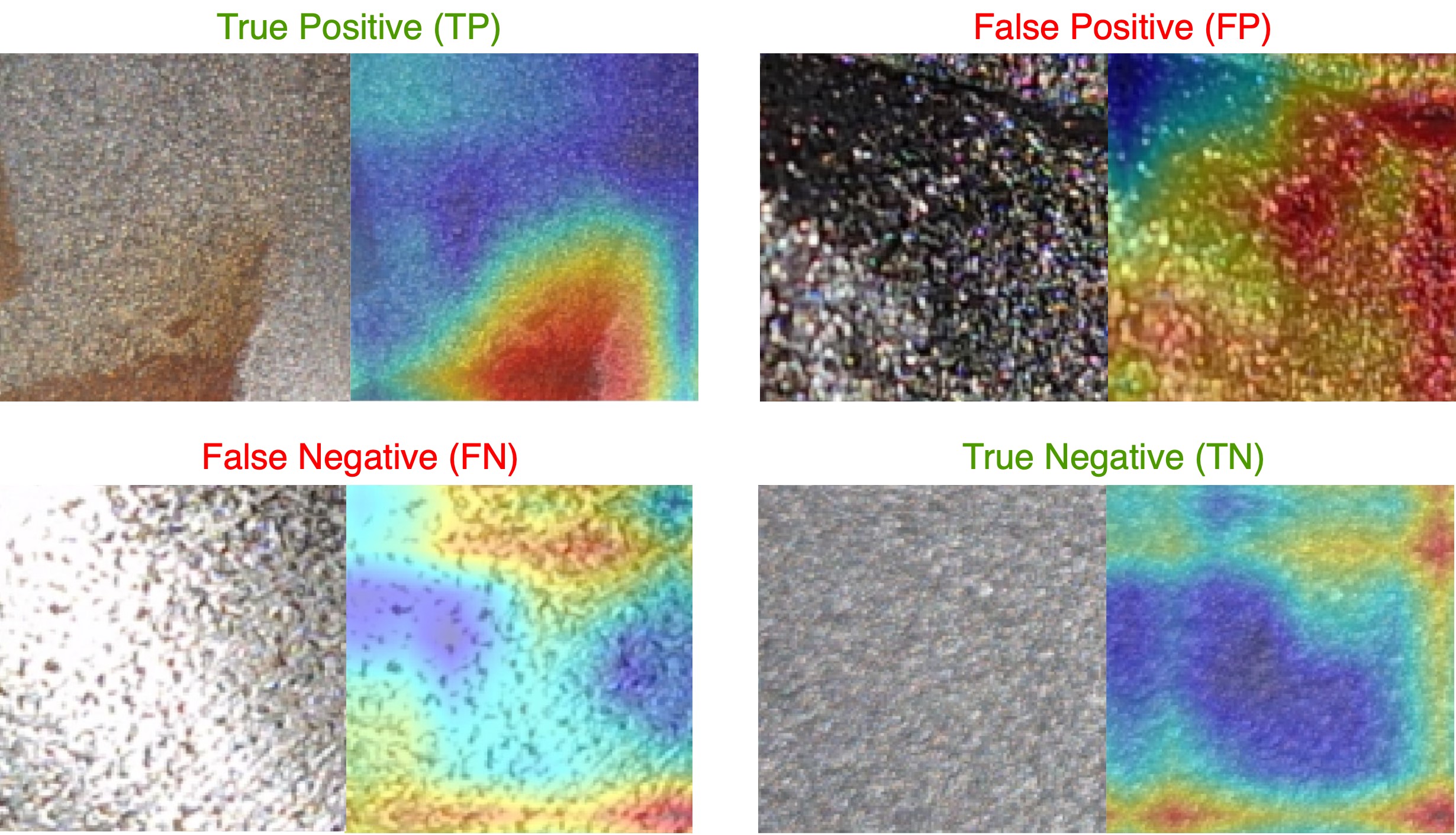}
    \caption{Interpretation heatmap examples of truly and falsely classified images using CCT}
    \label{fig:cct-cm}
\end{figure}

We include several heatmaps (Figure~\ref{fig:cct-cm}) of the small CCT model to demonstrate the interpretability of the approach. We found that identified features remain consistent across different model sizes. In the true positive cases, the heatmaps correctly highlight defect regions, demonstrating that the model effectively detects anomalies. False positives, though relatively rare, reveal that the model sometimes misinterprets normal surface textures as defects, leading to occasional misclassifications. Similarly, false negatives are infrequent but occur when the model overlooks subtle defects that resemble normal areas. True negatives, on the other hand, confirm that the model successfully recognizes defect-free surfaces. This consistency indicates that, even without precise defect localization data, the model can accurately highlight relevant areas, providing valuable insights.

\begin{figure}[htbp]  
    \centering
    \includegraphics[width=0.25\textwidth]{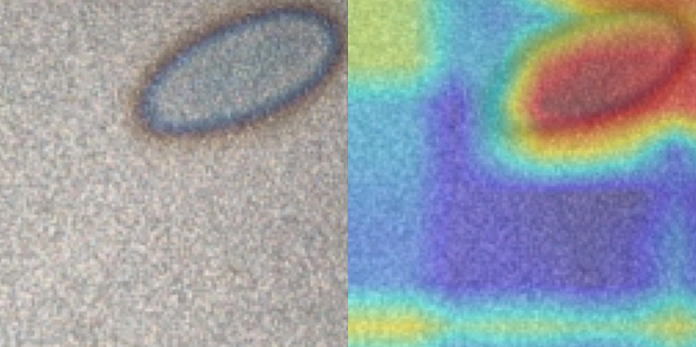}  
    \caption{CCT model heatmap on a defective surface}
    \label{fig:single CCT heatmap} 
\end{figure}

A particularly interesting case is shown in the additional heatmap (Figure \ref{fig:single CCT heatmap}), where a distinct elliptical spot is clearly marked as an anomaly. The model assigns high activation values to this region, demonstrating its ability to capture subtle surface irregularities. However, while the model performs well on well-defined defects, it occasionally struggles with ambiguous textures and lighting variations, which can lead to rare false positives or missed defects. The overall low frequency of these errors suggests that the approach is generally robust, making it a promising tool for real-world defect detection.

\subsection{Convolutional Autoencoders (CAE)}

\subsubsection{Experimental Setup}
The input images were resized to $128\times128$ pixels, with no further processing. Only the "ready for paint" samples were used for autoencoder training. The model was trained for 50 epochs, using the Adam optimizer with a learning rate of $0.001$ and a batch size of $32$. The weights were saved after every step in order to afterwards determine the optimal degree of autoencoder training. The implementation, training and evaluation were completed in Python and PyTorch.

\subsubsection{Results}
By the autoencoder's nature, with longer training it learns to reconstruct the input better. For the application at hand, there is a trade-off between reconstruction accuracy and the ease to differentiate between normal and defective inputs. Fewer epochs do not allow learning enough meaningful features, while training for too long leads to overly accurate replications of anomalous inputs. With the chosen architecture and thresholding parameter, the best result was yielded after 20 epochs, showing 67\% accuracy, 61\% precision, 77\% recall and F1 score of 0.68. The relatively high recall suggests effective defect identification.

\subsubsection{Heatmaps}
\begin{figure}
    \centering
    \includegraphics[width=0.5\linewidth]{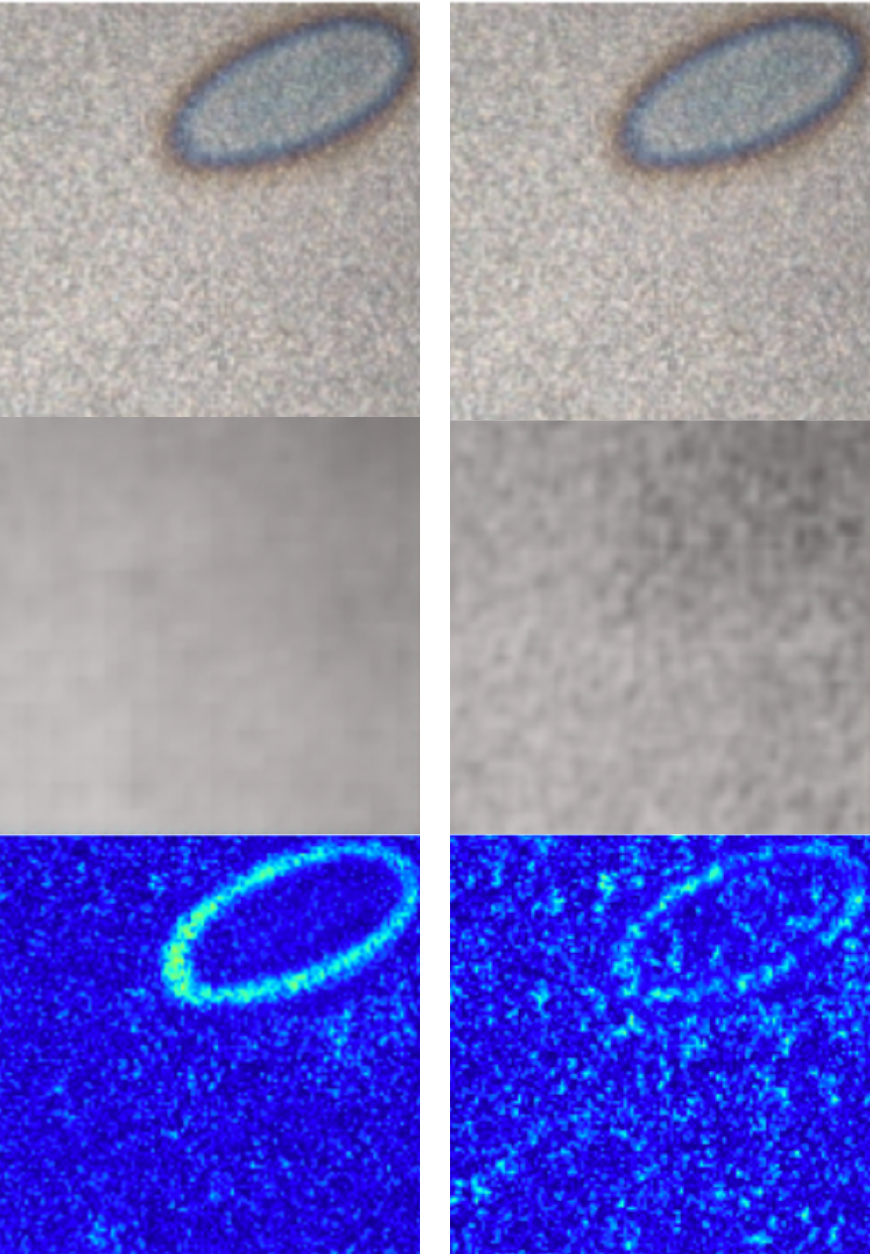}
    \caption{Original image (top), reconstructed image (middle), error heatmap
(bottom) after 20 epochs (left) and 50 epochs (right).}
    \label{fig:cae_hmap}
\end{figure}

Since the classification is based solely on the reconstruction error, this is what we visualized in this section to further support the capabilities of this approach. As expected, the selected model captures the general representation of the training data, producing what looks like a strongly blurred copy of the original. When presented with an anomalous image, the CAE struggles to reconstruct the defect, exactly as intended. This is clearly shown on the left in Figure \ref{fig:cae_hmap}: the ellipse (an example of discoloration) is not present in the output image, resulting in a matching shape appearing in a brighter colour on the heatmap. In the right half of the same figure is the visualization after 50 epochs. We can see a more defined dark spot in the reconstruction where the defect is located in the original image, showing signs of overfitting. 

\begin{figure}
    \centering
    \includegraphics[width=0.9\linewidth]{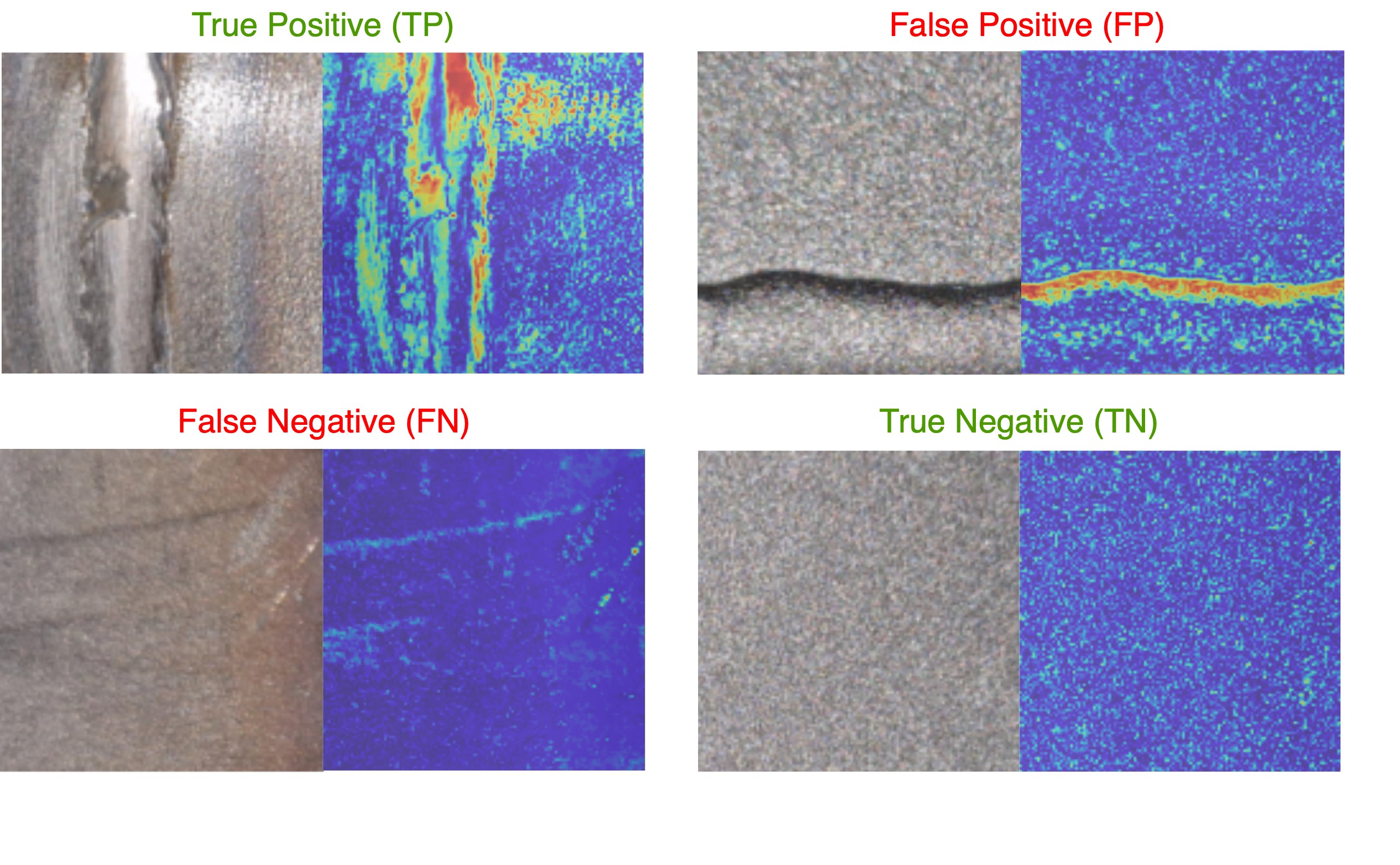}
    \caption{Interpretation heatmap examples of truly and falsely classified images using CAE}
    \label{fig:cae-cm}
\end{figure}

A few more examples are shown in Figure \ref{fig:cae-cm}. Our approach works well on smooth "ready for paint" surfaces and correctly detects untreated welds and abrasion. However, the CAE is easily confused by shadows, which leads to images such as the top-right one being misclassified as "needing shot-blasting". Additionally, defective inputs showing texture similar to the desired one often add to false negatives.

\subsection{Comparison of Methods}

The benchmarking results of the three methods on our SteelBlastQC dataset are shown in Table~\ref{tab:method_comparison}. CCT and SVM with ResNet50 feature extraction both achieved $95.0$\% accuracy with very similar precision, recall, and F1-scores of $0.95$. While CCT leverages transformer-based attention for end-to-end learning, SVM relies on pre-trained ResNet-50 features, making it less computationally demanding but dependent on feature quality. Heatmap analysis showed that while SVM’s grayscale heatmaps improved interpretability, CCT’s heatmaps were more accurate across all configurations, consistently highlighting defect regions. Both models perform well, with CCT offering a fully trainable deep learning approach and SVM benefiting from efficient pre-trained feature extraction, making the choice dependent on resource constraints and interpretability needs.

The CAE-based approach is a weak competitor against the other two methods, scoring around 0.3 lower on all four considered metrics. However, we can still see that in general it is a viable option, producing stable and explainable results without the need for labeled data or external feature extraction models.

\begin{table}[htbp]
\centering
\renewcommand{\arraystretch}{1.5} 
\setlength{\tabcolsep}{9pt} 
\small
\caption{Benchmarking results comparison on SteelBlastQC dataset}
\begin{tabular}{l c c c c}
\hline
\textbf{Method} & \textbf{Accuracy} & \textbf{Precision} & \textbf{Recall} & \textbf{F1-score} \\ \hline
CCT & 0.950 & 0.950 & 0.950 & 0.950 \\ \hline
SVM & 0.945 & 0.945 & 0.955 & 0.950 \\ \hline
CAE & 0.676 & 0.610 & 0.768 & 0.680 \\ \hline
\end{tabular}
\vspace{10pt} 
\label{tab:method_comparison}
\end{table}

\section{Conclusion and Discussion}

This study evaluates the potential of deep learning models for automating the quality control of shot-blasted steel surfaces. The key contribution of this work is the creation of the SteelBlastQC dataset, specifically tailored for this task. The dataset contains images of metal surfaces classified as either "ready for paint" or "needing shot-blasting." Three different supervised and unsupervised methods were adopted to validate the usability of the dataset. Supervised learning with CCT and SVM with ResNet-50 feature extraction achieved the best performance. CCT effectively captured both fine-grained texture details and broader surface patterns, making it well-suited for this classification task. However, its larger configurations require significant computational resources. SVM provided a computationally simpler and more interpretable alternative, with RGB images yielding higher accuracy, while grayscale models offered enhanced feature interpretability.

Further improvements could be achieved through improving input image resolution and designing dedicated network architectures. SVM with ResNet-50 feature extraction could similarly benefit from training on higher-resolution images and data augmentation to improve feature robustness. Exploring recent advances in deep learning research, such as Swin Transformers \cite{liu2021swin} and Feature Pyramid Networks \cite{min2022attentional} for hierarchical feature learning, can inspire the design of dedicated architectures and improve the baseline performance shown in this work.

As for exploring unsupervised learning techniques, our results validate the potential of an autoencoder-based approach for shot-blasting quality control. However, our approach uses only half of the dataset without defects for model training. Further exploration of advances in self-supervised learning, including contrastive learning methods (like SimCLR \cite{chen2020simple}, MoCo \cite{he2020momentum}), can learn robust representations from both positive and negative data and improve generalization. Additionally, this approach can be combined with few-shot learning methods to adapt to new data with new types of defects.

Moreover, heatmaps provide valuable insights into each model's behavior, revealing how models identify surface defects such as welding lines and texture inconsistencies. This can be expanded to defect localization, which will save the high cost of labeling accurate defect areas.

Overall, this study demonstrates that automated defect detection using computer vision is a viable alternative to manual inspection. Automated vision systems powered by the CCT, SVM, or CAE methods could reduce human error, lower inspection costs, and streamline production workflows. Future work should focus on refining models for deployment in industrial environments and optimizing computational efficiency. By making our dataset and code openly available, we aim to facilitate further advancements in automated quality control and encourage future research in industrial defect detection.

\bibliographystyle{IEEEtran}
\bibliography{ref}

\begin{appendices}

\begin{figure*}[htbp]  
    \centering
    \includegraphics[width=1\textwidth]{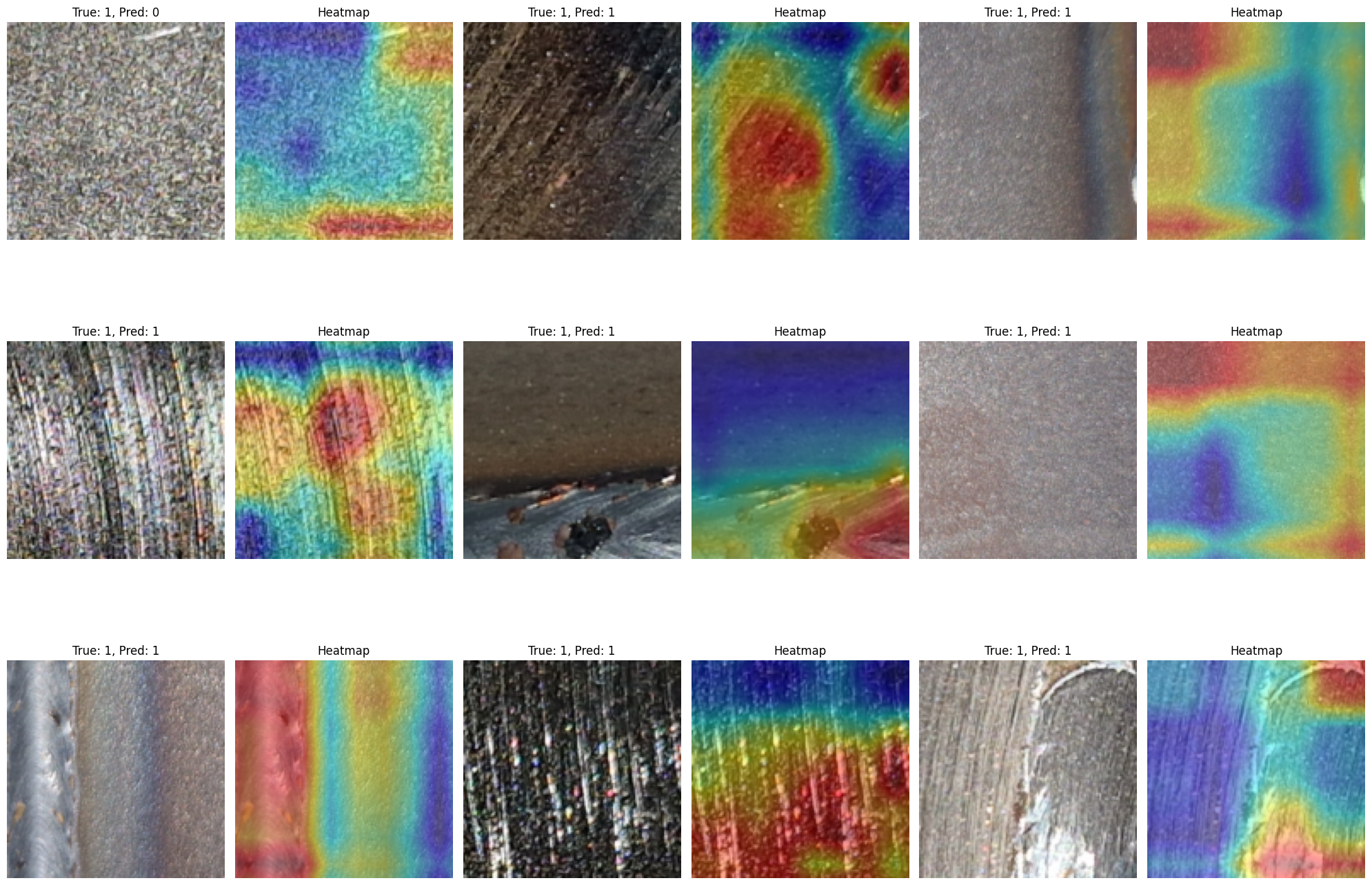}  
    \caption{Heatmaps of defective surfaces for the CCT small model.}
    \label{cctheatmaps}
\end{figure*}

\begin{figure*}[htbp]  
    \centering
    \includegraphics[width=1\textwidth]{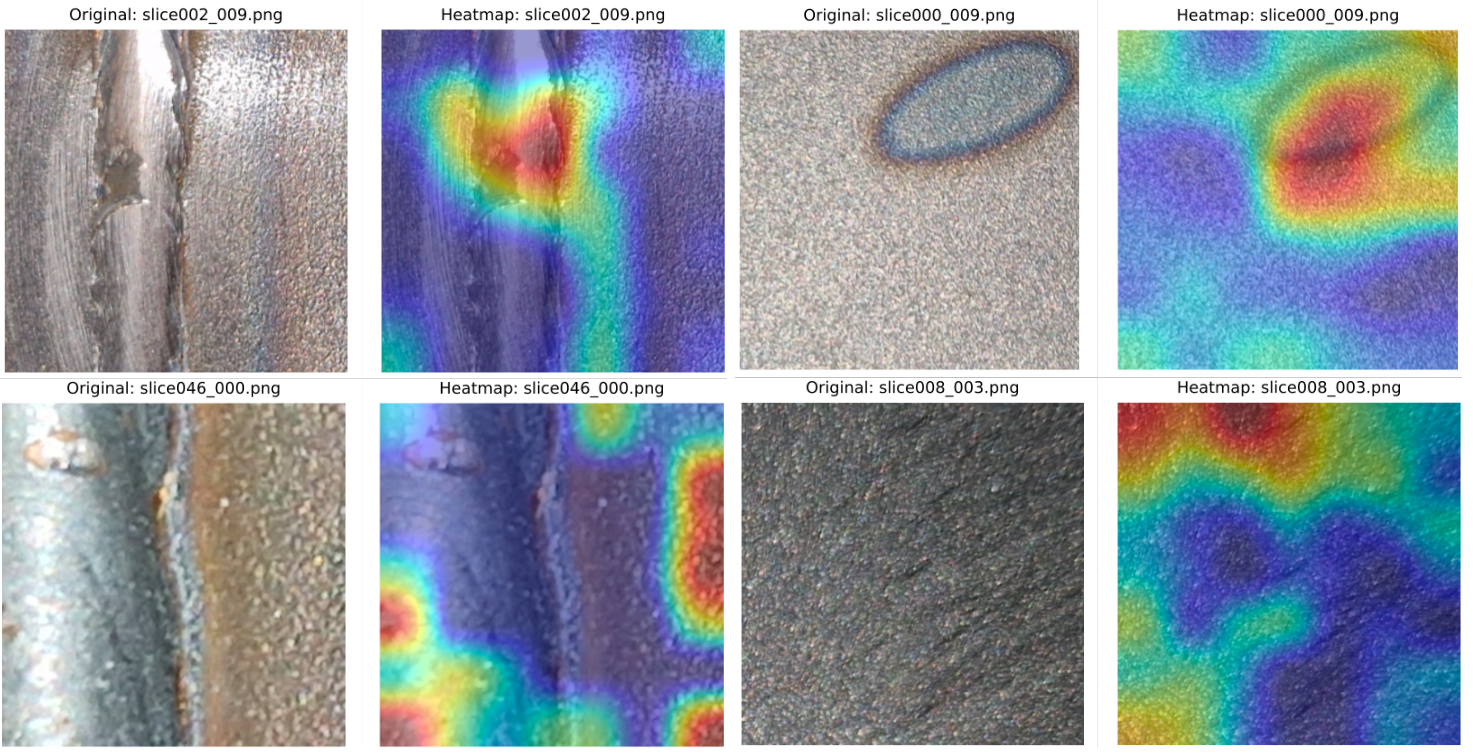}  
    \caption{Heatmaps of defective surfaces for the SVM with ResNet50 Feature Extraction method.}
    \label{SVMheatmaps}
\end{figure*}

\section{Heatmaps (Figure~\ref{cctheatmaps} and Figure~\ref{SVMheatmaps})}

We provide additional heatmaps for the CCT small model and SVM with ResNet-50 feature extraction. CCT heatmaps exhibit more precise attention, effectively localizing surface defects such as welding seams and rough textures. In contrast, SVM heatmaps show broader activation, sometimes lacking refined focus. While both models perform well, CCT demonstrates superior defect localization. 

\end{appendices}

\end{document}